\documentclass[11pt]{article}
\pdfoutput=1

\usepackage[T1]{fontenc}              % T1 font encoding as default
\usepackage[latin9]{inputenc}         % ISO-8859-15 (Latin 9) text

\usepackage[a4paper]{geometry}        % DIN A4 paper
\usepackage{mathpazo}
\usepackage{microtype}
\usepackage{natbib}

\usepackage{url}
\usepackage{booktabs}
\usepackage{graphicx} 
\usepackage{amssymb,amsfonts,amsmath}

\newcommand{\figcite}[1]{Fig.~\textbf{\ref{#1}}}
\newcommand{\eqcite}[1]{Eq.~\textbf{\ref{#1}}}
\newcommand{\tabcite}[1]{Tab.~\textbf{\ref{#1}}}

\newcommand\bx{\boldsymbol x}
\newcommand\bI{\boldsymbol I}
\newcommand\bmu{\boldsymbol \mu}

\newcommand\tml{\hat\theta_k^\text{ML}}

\newcommand\btheta{\boldsymbol\theta}
\newcommand\hbtheta{\hat\btheta}

\begin{document}

\date{5 October 2008; last revised 19 June 2009}

\title{Entropy inference and the James-Stein estimator, 
       with application to nonlinear gene association networks}

\author{Jean Hausser \thanks{Email: jean.hausser@unibas.ch; Address:
         Bioinformatics, Biozentrum,
        University of Basel,
       Klingelbergstr. 50/70,
       CH-4056 Basel, Switzerland}
\ and
 Korbinian Strimmer \thanks{Email: strimmer@uni-leipzig.de; Address:
         Institute for Medical Informatics,
      Statistics and Epidemiology,
      University of Leipzig,
      H\"artelstr. 16--18,
      D-04107 Leipzig, Germany }
}

\maketitle

\begin{abstract}
We present a procedure for effective estimation of entropy and mutual
information from small-sample data, and apply it to the problem of
inferring high-dimensional gene association networks. Specifically, we
develop a James-Stein-type shrinkage estimator, resulting in a procedure
that is highly efficient statistically as well as computationally.
Despite its simplicity, we show that it outperforms eight other entropy
estimation procedures across a diverse range of sampling scenarios and
data-generating models, even in cases of severe undersampling. We
illustrate the approach by analyzing \emph{E.~coli} gene expression data and
computing an entropy-based gene-association network from gene expression
data. A~computer program is available that implements the proposed
shrinkage estimator.
\end{abstract}


\vspace{1cm}

\noindent{\bf Acknowledgments}
This work was partially supported by 
an Emmy Noether grant of the Deutsche Forschungsgemeinschaft (to K.S.).
We thank the anonymous referees and the editor for very helpful comments.

\newpage
\section{Introduction}

Entropy is a fundamental quantity in statistics and machine learning.
 It has a large number of applications, for example in 
astronomy, cryptography, signal processing, statistics, physics, image analysis 
neuroscience, network theory, and bioinformatics---see, 
for example, \citet{Stin06}, \citet{YB04}, \citet{Mac03} and \citet{SK+98}.  Here we focus on \emph{estimating}
entropy from small-sample data, with applications in genomics and gene network
inference in mind \citep{MNB+06,MKLB07}.

To define the Shannon entropy, consider a categorical random variable 
with alphabet size $p$
and associated cell probabilities $\theta_1, \dots, \theta_p$ with 
$\theta_k > 0$ and $\sum_k \theta_k = 1$.  Throughout the article, we assume that
$p$ is fixed and known.
In this setting, the Shannon entropy in natural units is given by\footnote{In this
paper we use the following conventions:  $\log$ denotes the natural logarithm (\emph{not} base 2 or base 10),
and we define $0 \log 0 = 0$.}  
\begin{equation}
H = - \sum_{k=1}^p \theta_k \log( \theta_k) .
\label{eq:shannon}
\end{equation}
In practice, the underlying probability mass function are unknown, hence 
 $H$ and $\theta_k$ need to be \emph{estimated} from observed cell counts $y_k \geq 0$.

A particularly simple and widely used estimator of entropy is the maximum 
likelihood (ML) estimator
$$
\hat{H}^\text{ML} = - \sum_{k=1}^p \tml \log( \tml )
$$
constructed by plugging  the ML frequency estimates 
\begin{equation}
\tml = \frac{y_k}{n}
\label{eq:theta.ml}
\end{equation}
into \eqcite{eq:shannon}, with $n = \sum_{k=1}^p y_k$ being the total number of counts.

In situations with $n \gg p$, that is, when the dimension is low and when there are
many observation, it is easy to infer entropy reliably, and it is well-known that 
in this case the 
ML estimator is optimal.  However, in high-dimensional problems 
with $n \ll p$
it becomes extremely challenging to estimate the entropy.
Specifically, in the ``small $n$, large $p$'' regime the ML estimator
performs very poorly and severely underestimates the true entropy.

While entropy estimation has a long history tracing back to
more than 50 years ago, it is only recently that the specific issues
arising in high-dimensional, undersampled data sets have attracted attention.
This has lead to two recent innovations, namely the NSB algorithm 
\citep{NSB02} and the Chao-Shen  estimator \citep{CS03}, both
of which are now widely considered as benchmarks for the small-sample 
entropy estimation problem \citep{VYK07}.

Here, we introduce a novel and highly efficient small-sample entropy
estimator based on James-Stein shrinkage \citep{Gru98}.  Our method
is fully analytic and hence computationally inexpensive.
Moreover, our procedure simultaneously provides estimates of the entropy
\emph{and} of the cell frequencies suitable for plugging into the Shannon
entropy formula (\eqcite{eq:shannon}).  Thus, in comparison the estimator 
we propose is simpler, very efficient, and at the same time more versatile 
than currently available entropy estimators.

\section{Conventional Methods for Estimating Entropy}

Entropy estimators can be divided into two
groups: i) methods, that rely on estimates of %underlying
cell frequencies, and ii)~estimators, that directly infer entropy
without estimating a compatible set of $\theta_k$.  
Most methods discussed below fall into the first group, except for
the Miller-Madow and NSB approaches.

\subsection{Maximum Likelihood Estimate}
The connection between observed counts $y_k$ and frequencies $\theta_k$ 
is given by the multinomial distribution 
\begin{equation}
\label{eq:multinomial}
\text{Prob}(y_1, \hdots, y_p; \theta_1,\hdots, \theta_p) = 
\frac{n!}{\prod_{k=1}^p y_k! } \prod_{k=1}^p \theta_k^{y_k} .
\end{equation}
Note that $\theta_k >0 $ because otherwise  
the distribution is singular.
In contrast, there may be (and often are) zero counts $y_k$.  
The ML estimator of $\theta_k$ maximizes the right hand side of
\eqcite{eq:multinomial} for fixed $y_k$,
leading to the   observed  frequencies
$\tml = \frac{y_k}{n}$ with variances 
$\text{Var}(\tml) = \frac{1}{n} \theta_k (1-\theta_k)$
and $\text{Bias}(\tml) = 0$ as $E(\tml) = \theta_k$.

\subsection{Miller-Madow Estimator}
While $\tml$ is unbiased, the corresponding plugin entropy estimator 
$\hat{H}^\text{ML}$ is not. First order bias correction  leads to
$$
\hat{H}^\text{MM} = \hat{H}^\text{ML} + \frac{m_{>0}-1}{2 n},
$$
where $m_{>0}$ is the number of cells with $y_k > 0$.
This is known as the Miller-Madow  estimator \citep{Mil55}. 

\subsection{Bayesian Estimators}
Bayesian regularization of cell counts \emph{may} lead to vast improvements
over the ML estimator \citep{AH05}. Using the Dirichlet distribution 
with parameters $a_1, a_2, \hdots, a_p$ as prior, the resulting posterior  
distribution is also Dirichlet with mean
$$
\hat\theta_k^\text{Bayes} = \frac{y_k + a_k}{n + A} ,
$$
where $A = \sum_{k=1}^p a_k$. The flattening constants $a_k$  play the 
role of pseudo-counts (compare with \eqcite{eq:theta.ml}), so that  $A$ may be interpreted 
as the \emph{a priori} sample size.

\begin{table}[b]
%\caption{Common choices for the parameters of the Dirichlet prior  in
%the Bayesian estimators of cell frequencies, and corresponding entropy estimators.}
\centering
\begin{tabular}{rrr}
\toprule
$a_k$  & Cell frequency prior  & Entropy estimator \\
\midrule
$0$ & no prior & maximum likelihood \\
$1/2$ & Jeffreys prior \citep{Jef46} & \citet{KT81}  \\
$1$ & Bayes-Laplace uniform prior &  \citet{HGH98} \\
$1 / p$ & Perks prior \citep{Per47} &  \citet{SG96} \\
$ \sqrt{n}/p$ &  minimax prior \citep{Try58} & \\
\bottomrule
\end{tabular}
\caption{Common choices for the parameters of the Dirichlet prior  in
the Bayesian estimators of cell frequencies, and corresponding entropy estimators.}

\label{tab:bayes.est}
\end{table}

Some common choices for  $a_k$ are listed in \tabcite{tab:bayes.est},
\nocite{Jef46} 
\nocite{KT81}  
\nocite{HGH98}  
\nocite{Per47}   
\nocite{SG96}  
\nocite{Try58}  
along with references to the corresponding plugin entropy estimators, 
$$
\hat H^{\text{Bayes}} = - \sum_{k=1}^p \hat\theta_k^{\text{Bayes}} \log( \hat\theta_k^{\text{Bayes}}) .
$$
While the multinomial model with Dirichlet prior is standard 
Bayesian folklore \citep{GCSR04}, there is no general agreement regarding which 
assignment of $a_k$ is best as noninformative prior---see
for instance the discussion in \citet{TGM08} and \citet{Gei84}.
But, as shown later in this article, choosing inappropriate $a_k$ can easily cause the resulting estimator
to perform \emph{worse} than the ML estimator, thereby defeating the originally 
intended purpose.

\subsection{NSB Estimator}
The NSB approach \citep{NSB02} avoids overrelying on a particular choice of $a_k$ in the  Bayes 
estimator by using a more refined prior. Specifically, a Dirichlet mixture prior with infinite number
of components is employed, constructed such that the resulting prior over the entropy is uniform. 
While the NSB estimator is one of the best entropy estimators available at present in terms of statistical 
properties, using the Dirichlet mixture prior  is computationally expensive
and somewhat slow for practical applications.

\subsection{Chao-Shen Estimator}
Another recently proposed estimator is due to \citet{CS03}.
This approach applies the Horvitz-Thompson estimator
\citep{HT52} in combination with the Good-Turing correction \citep{Good53,OSZ03}
of the empirical cell probabilities
to the problem of entropy estimation.  The Good-Turing-corrected frequency estimates are
$$
\hat\theta_k^{\text{GT}} = (1-\frac{m_1}{n}) \tml , 
$$
where $m_1$ is the number of singletons, that is, cells with $y_k=1$.
Used jointly with the Horvitz-Thompson estimator this results in
$$
\hat H^{CS} = -\sum_{k = 1}^p \frac{\hat\theta_k^{\text{GT}} \log \hat\theta_k^{\text{GT}}}{(1-(1-\hat\theta_k^{\text{GT}})^n)},
$$ 
an estimator with remarkably good statistical properties \citep{VYK07}.  

%%% deleted on request of one referee
%Note that the Chao-Shen method relies on corrected estimates of cell frequencies
%that are not proper (i.e., by construction for $m_1 > 0$ the $\hat\theta_k^{\text{GT}}$ 
%don't sum up to 1). Hence, it is not possible to use them in conjunction with the 
%Shannon formula (\eqcite{eq:shannon}).

\section{A James-Stein Shrinkage Estimator}

The contribution of this paper is to introduce an entropy estimator
that employs James-Stein-type shrinkage at the level of cell frequencies.
As we will show below, this  leads to an entropy estimator that is highly effective,
both in terms of statistical accuracy and computational complexity.

James-Stein-type shrinkage is a simple analytic device to perform regularized 
high-dimensional inference.  It is ideally suited for small-sample settings - 
the original estimator \citep{JS61} considered sample size $n=1$.  
A general recipe for constructing shrinkage estimators
is given in \emph{Appendix A}. In this section, we describe how this approach can 
be applied to the specific problem of  estimating cell frequencies.

James-Stein shrinkage is based on averaging two very different models: 
a high-dimensional model with low bias and high variance, and a lower 
dimensional model with larger bias but smaller variance. The intensity of the 
regularization is determined by the relative weighting of the two models.
Here we consider the convex combination
\begin{equation}
\label{eq:shrinktheta}
\hat\theta_k^\text{Shrink} = \lambda t_k + (1-\lambda) \tml ,  
\end{equation}
where $\lambda \in [0,1]$ is the shrinkage intensity 
that takes on a value between 0 (no shrinkage) and 1 (full shrinkage), and
$t_k$ is the shrinkage target. A convenient choice of $t_k$ is
the uniform distribution $t_k = \frac{1}{p}$.  This is also the  
maximum entropy target.   Considering that  $\text{Bias}(\tml) = 0$ and using
 the unbiased estimator  $\widehat{\text{Var}}(\tml) = \frac{\tml (1-\tml)}{n-1}$
we obtain (cf. \emph{Appendix A}) for the shrinkage intensity 
\begin{equation}
\hat{\lambda}^\star = 
        \frac{\sum^p_{k=1} \widehat{\text{Var}}(\tml)}{
                    \sum^p_{k=1} (t_k - \tml)^2} = 
         \frac{ 1- \sum^p_{k=1} (\tml)^2}{(n-1)
                    \sum^p_{k=1} (t_k - \tml)^2}.
\label{eq:opt.lambda.hat}
\end{equation}
Note that this also assumes a non-stochastic target $t_k$.
The resulting plugin shrinkage entropy estimate is 
\begin{equation}
\label{eq:shrinkentropy}
\hat H^{\text{Shrink}} = - \sum_{k=1}^p \hat\theta_k^{\text{Shrink}} \log( \hat\theta_k^{\text{Shrink}}) .
\end{equation}

\subsubsection*{Remark 1:}
There is a one to one correspondence between the shrinkage and the Bayes estimator. If we write 
 $t_k = \frac{a_k}{A}$ and $\lambda = \frac{A}{n+A}$, 
 then $ \hat\theta_k^\text{Shrink} = \hat\theta_k^\text{Bayes}$.
This implies that the shrinkage estimator is an empirical Bayes estimator
with a data-driven choice of the flattening constants---see also \citet{EM73}.
For every choice of $A$ there exists an equivalent shrinkage intensity $\lambda$.
Conversely,
for every $\lambda$ there exist an equivalent $A = n \frac{\lambda}{1-\lambda}$.

\subsubsection*{Remark 2:}
Developing $A = n \frac{\lambda}{1-\lambda} = n (\lambda + \lambda^2 + \ldots)$ 
we obtain the approximate estimate $\hat A = n \hat \lambda$, which in turn recovers
 the ``pseudo-Bayes'' estimator described in \citet{FH73}.

\subsubsection*{Remark 3:}
The shrinkage estimator assumes a fixed and known $p$.   In many practical applications
this will indeed be the case, for example, if the observed counts are due to discretization (see also the 
data example).  In addition, the shrinkage estimator appears to be robust against 
assuming a larger $p$ than necessary  (see scenario 3 in the simulations).

\subsubsection*{Remark 4:}
The shrinkage approach can easily be modified to allow multiple targets with different shrinkage intensities. 
For instance, using the Good-Turing estimator \citep{Good53,OSZ03}, 
one could setup a different uniform target for the non-zero and the zero counts, respectively.

\section{Comparative Evaluation of Statistical Properties}

In order to elucidate the relative strengths and weaknesses of the entropy estimators
reviewed in the previous section, we set to benchmark them in a simulation 
study covering different data generation processes and sampling regimes. 

\subsection{Simulation Setup}
We compared the statistical performance of all nine described estimators 
(maximum likelihood, Miller-Madow, four Bayesian estimators, 
the proposed shrinkage estimator (Eqs.~\textbf{\ref{eq:shrinktheta}}--\textbf{\ref{eq:shrinkentropy}}), 
NSB und Chao-Shen) under various sampling
and data generating scenarios:
\begin{itemize}
\item The dimension was fixed at $p=1000$.
\item Samples size $n$ varied from $10$, $30$, $100$, $300$, $1000$, $3000$, 
      to $10000$.  That is, we investigate cases of dramatic undersampling (``small $n$, large $p$'') 
      as well as situations with a larger number of observed counts.
\end{itemize}
The true cell probabilities $\theta_1, \dots, \theta_{1000}$ were assigned in four different fashions, corresponding to
 rows 1-4 in \figcite{fig:simresults}: 
\begin{enumerate}
 \item Sparse and heterogeneous, following a Dirichlet distribution with parameter $a=0.0007$,
\item Random and homogeneous, following a Dirichlet distribution with parameter $a=1$,
\item As in scenario 2, but with half of the cells containing structural zeros, and
\item Following a Zipf-type power law. 
\end{enumerate}
For each sampling scenario and sample size, we conducted 1000 simulation runs.
In each run, we generated a new set of true cell frequencies and subsequently
sampled observed counts $y_k$ from the corresponding multinomial distribution.  The
resulting counts $y_k$ were then supplied to the various entropy and cell frequencies estimators
and the squared error $\sum_{i=k}^{1000}(\theta_k - \hat\theta_k)^2$ 
was computed.
From the 1000 repetitions we estimated the mean squared error (MSE) of the cell frequencies
by averaging over the individual squared errors (except for the NSB, Miller-Madow, and 
Chao-Shen estimators). Similarly, we computed estimates of MSE and bias of the inferred entropies.

\begin{figure*}[!pht]
\begin{center}
\centerline{\includegraphics[width=1\textwidth]{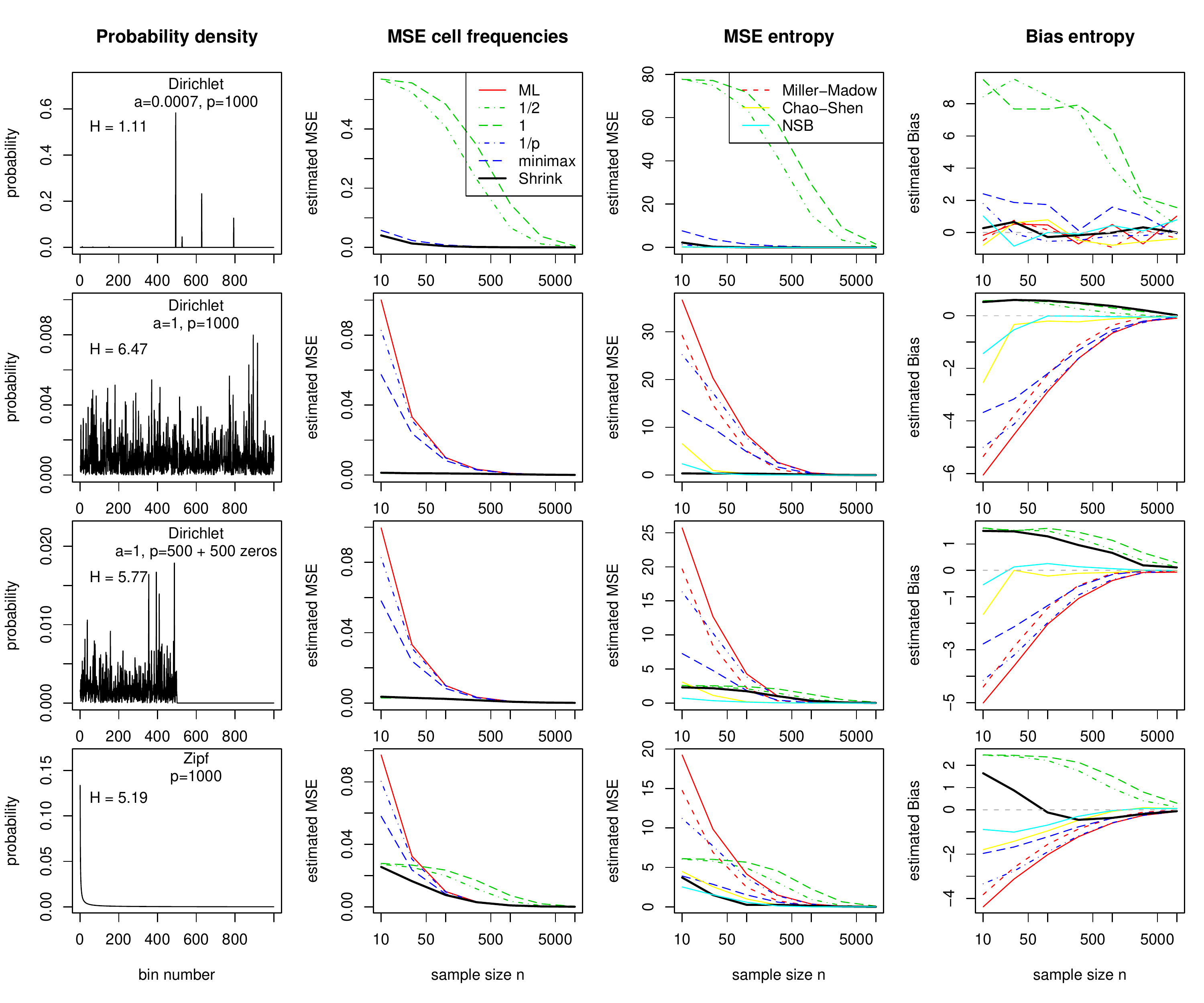}}
\caption{Comparing the performance of nine different entropy estimators (maximum
likelihood, Miller-Madow, four Bayesian estimators, the proposed shrinkage estimator, 
NSB und Chao-Shen) in four different sampling scenarios (rows 1 to 4).  The estimators are
compared in terms of MSE of the underlying cell frequencies (except for Miller-Madow, NSB, Chao-Shen)
and according to MSE and Bias of the estimated entropies.  The
dimension is fixed at $p=1000$ while the sample size $n$ varies from 10 to 10000.}
\label{fig:simresults}
\end{center}
\end{figure*}

\subsection{Summary of Results from Simulations}

\figcite{fig:simresults} displays the results of the simulation study, which can be summarized
as follows:

\begin{itemize}
\item Unsurprisingly, all estimators perform well when the sample size is large.
\item The maximum likelihood and Miller-Madow estimators perform worst, 
except for scenario 1. Note that these estimators are inappropriate even
for moderately large sample sizes. Furthermore, the bias correction 
of the Miller-Madow estimator is not particularly effective.
\item The minimax and $1/p$ Bayesian estimators tend to perform slightly better than
maximum likelihood, but not by much.
\item The Bayesian estimators with pseudocounts $1/2$ and $1$ perform very well even for
small sample sizes in the scenarios 2 and 3. However, they are less efficient 
in scenario 4, and completely fail in scenario 1.  
\item Hence, the Bayesian estimators can perform better or worse than the ML estimator, 
depending on the choice of the prior and on the sampling scenario.
\item The NSB, the Chao-Shen and the shrinkage estimator all are statistically
 very efficient with small MSEs in all four scenarios, regardless of sample size. 
\item The NSB and Chao-Shen estimators are nearly unbiased in scenario 3.
\end{itemize}
The three top-performing estimators are the NSB, the Chao-Shen and the 
prosed shrinkage estimator. When it comes to estimating the entropy, these estimators can be considered
identical for practical purposes.  However, the shrinkage estimator is the only one
that simultaneously estimates cell frequencies suitable for use with the Shannon entropy formula (\eqcite{eq:shannon}), and it does so with high accuracy even for small samples.
In comparison, the NSB estimator is by far the slowest method: in our simulations, the shrinkage estimator 
was faster by a factor of 1000.

\section{Application to Statistical Learning of Nonlinear Gene Association Networks}

In this section we illustrate how the shrinkage entropy estimator can be applied to the problem
of inferring regulatory interactions between genes through estimating the nonlinear association network. 

\subsection{From Linear to Nonlinear Gene Association Networks}

One of the aims of systems biology is to understand the interactions among genes and their products
underlying the molecular mechanisms of cellular function as well as how disrupting these interactions may
lead to different pathologies. To this end, an extensive literature on the problem of gene regulatory network 
``reverse engineering'' has developed in the past decade \citep{Fri04}.
Starting from gene expression or proteomics data, different statistical learning procedures have been proposed 
to infer associations and dependencies among genes. Among many others, methods have been proposed 
to enable the inference of large-scale correlation networks \citep{BTS+00} and of high-dimensional partial 
correlation graphs \citep{DHJ+04,SS05a,MB06}, for learning vector-autoregressive \citep{OS07a} and state 
space models \citep{RA+04,LS08}, and to reconstruct directed ``causal'' interaction graphs \citep{KB07,OS07c}.

The restriction to linear models in most of the literature is 
owed at least  in part  to the already substantial challenges involved in
estimating linear high-dimensional dependency structures.  However, cell biology offers numerous 
examples of threshold and saturation effects, suggesting that linear models may not be sufficient to model 
gene regulation and gene-gene interactions.
In order to relax the linearity assumption and to capture nonlinear associations
among genes, entropy-based network modeling was recently proposed in the form of the 
ARACNE \citep{MNB+06} and MRNET \citep{MKLB07} algorithms.

The starting point of these two methods is to compute the  mutual information
$\text{MI}(X, Y)$ for all pairs of genes $X$ and $Y$, where $X$ and $Y$ represent the expression 
levels of the two genes for instance.  The mutual information
is the Kullback-Leibler distance from the joint probability density to
the product of the marginal probability densities:
\begin{equation}
\text{MI}(X, Y) = E_{f(x,y)} \left\lbrace \log \frac{  f(x,y)}{f(x) f(y)} \right\rbrace .
\label{eq:kullbackmi}
\end{equation}
The mutual information (MI) is always non-negative, symmetric, and equals zero 
only if $X$ and $Y$ are independent.  For normally distributed variables the
mutual information is closely related to the usual Pearson correlation,
$$
\text{MI}(X, Y) =  -\frac{1}{2} \log(1-\rho^2) .
$$
Therefore, mutual information is a natural measure of the association between genes,
regardless whether linear or nonlinear in nature.

\subsection{Estimation of Mutual Information}

To construct an entropy network, we first need to  
 \emph{estimate} mutual information for all pairs of genes.
The entropy  representation
\begin{equation}
\text{MI}(X,Y)=H(X)+H(Y)-H(X,Y) ,
\label{eq:midef}
\end{equation}
shows that MI can be computed from the joint and marginal entropies
of the two genes $X$ and $Y$.  Note that this definition is equivalent to
the one given in \eqcite{eq:kullbackmi} which is
based on the Kullback-Leibler divergence.
From \eqcite{eq:midef} it is also evident
 that $\text{MI}(X,Y)$ is the information shared between the
two variables.  

For gene expression data the estimation of MI and the underlying entropies
is challenging due to the small sample size, which requires the use of
a regularized entropy estimator such as the shrinkage approach we propose here.
Specifically, we proceed as follows:
\begin{itemize}
\item As a prerequisite the data must be discrete, with each measurement assuming
one of $K$  levels.  If the data are not already discretized, we propose 
employing the simple algorithm of \citet{FD81}, considering the measurements of all genes simultaneously. 
\item Next, we estimate the $p= K^2$ cell frequencies  of the $K \times K$ contingency 
table for each pair $X$ and $Y$ using the shrinkage approach (Eqs.~\ref{eq:shrinktheta} and ~\ref{eq:opt.lambda.hat}).
Note that typically the sample size $n$ is much smaller than $K^2$, thus simple
approaches such as ML are not valid.
\item Finally, from the estimated cell frequencies we calculate
$H(X)$, $H(Y)$, $H(X,Y)$ and the desired $\text{MI}(X,Y)$.
\end{itemize}

\subsection{Mutual Information Network for \emph{E. Coli} Stress Response Data}

\begin{figure*}[!ht]
\begin{center}
\centerline{\includegraphics[width=1\textwidth]{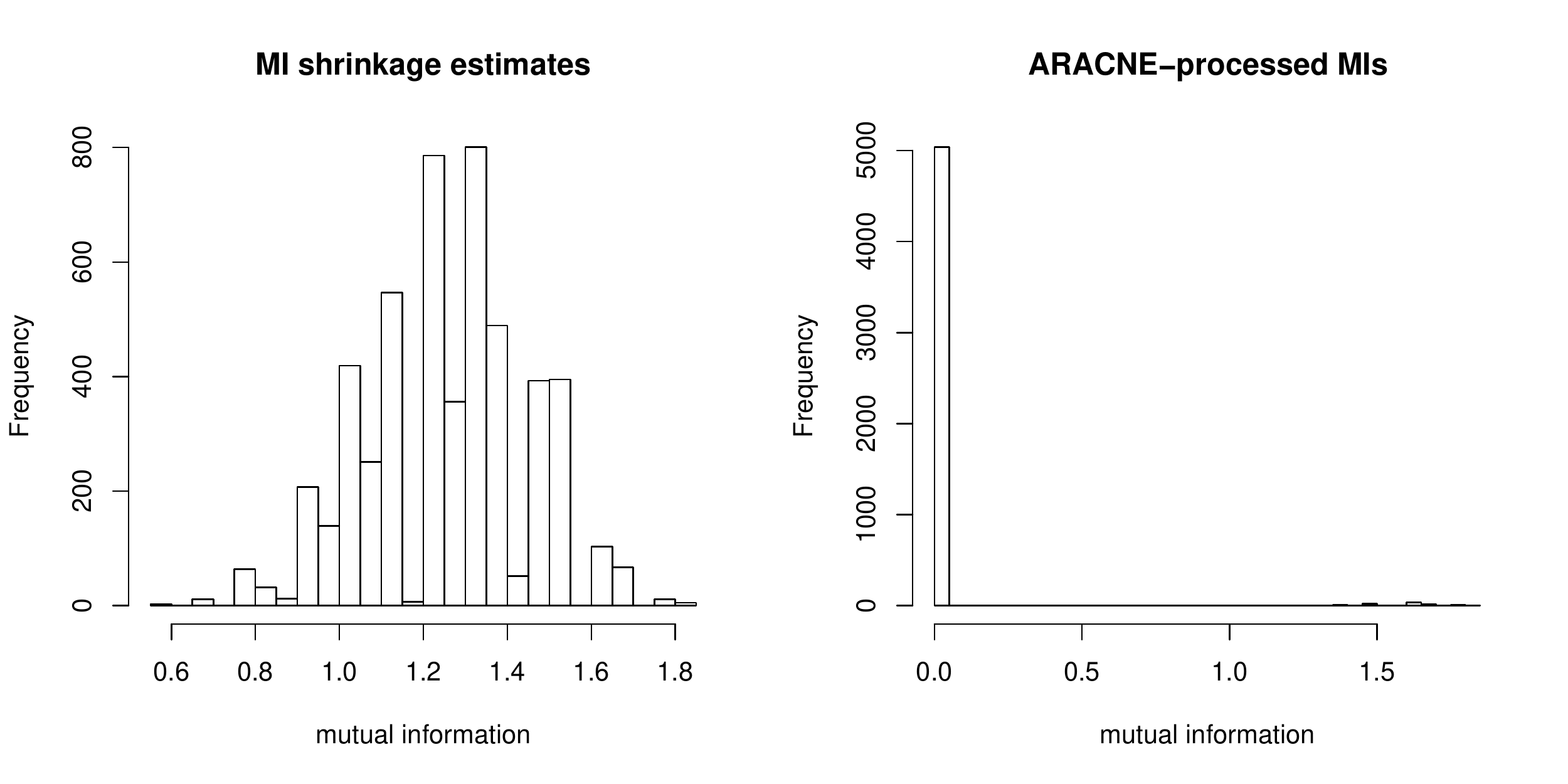}}
\caption{Left: Distribution of estimated mutual information values for all 5151 gene pairs
of the \emph{E. coli} data set. Right: Mutual information values after applying the ARACNE gene pair 
selection procedure.
 Note that the most MIs have been set to zero by the ARACNE algorithm.}
\label{fig:mihist}
\end{center}
\end{figure*}

\begin{figure*}[!p]
\begin{center}
\centerline{\includegraphics[width=1\textwidth]{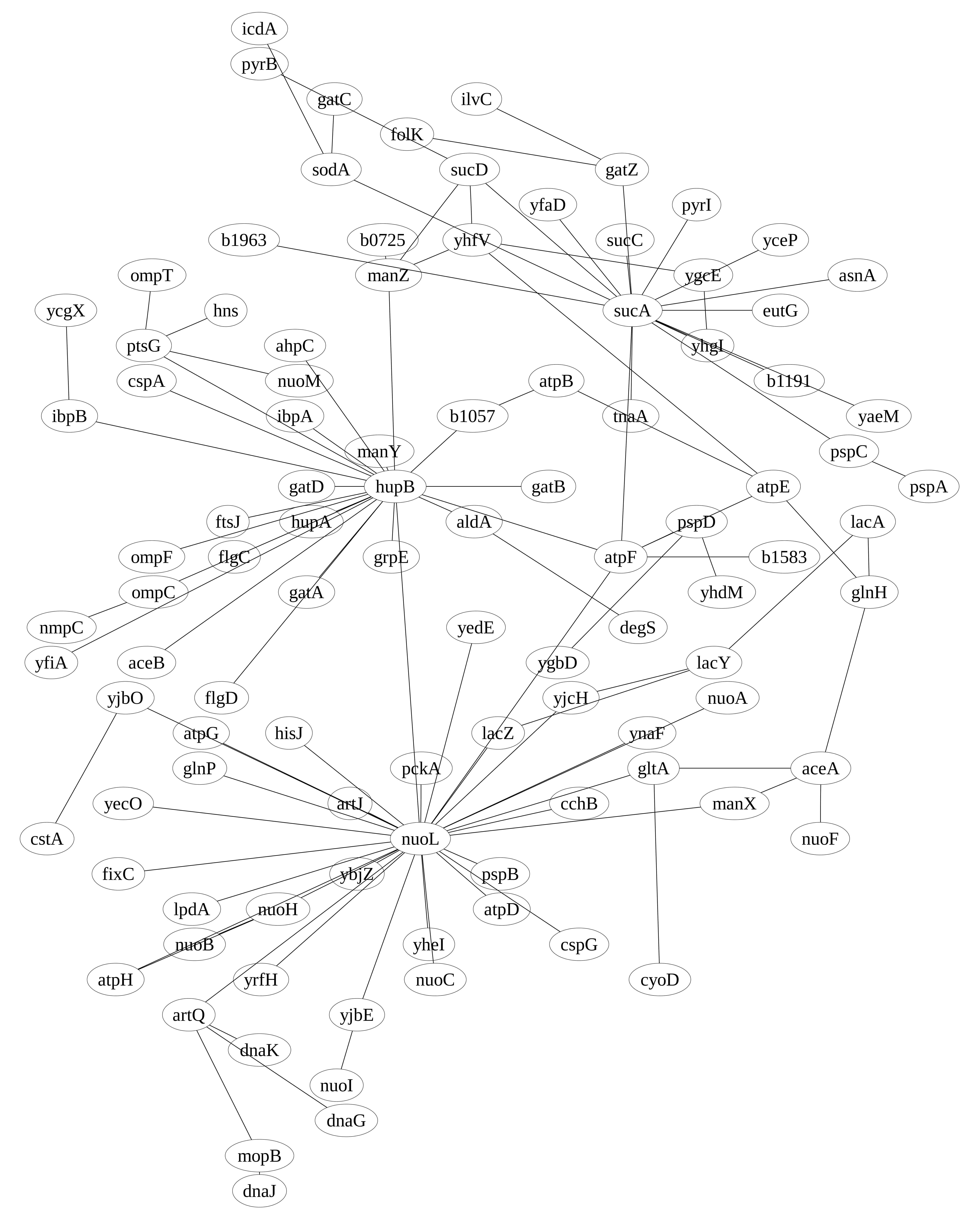}}
\caption{Mutual information network for the \emph{E. coli} data inferred by the
ARACNE algorithm based on shrinkage estimates of entropy and mutual information.}
\label{fig:ecolinet}
\end{center}
\end{figure*}

For illustration, we now analyze data from \citet{SGT+04}
who conducted an experiment to observe the 
stress response in \emph{E. Coli} during expression of a recombinant protein.
This data set was also used in previous linear network analyzes, for example, in \citet{SS05c}. 
The raw data consist of 4289 protein coding genes, on which measurements were taken
at 0, 8, 15, 22, 45, 68, 90, 150, and 180 minutes.  We focus on a subset
of $G=102$ differentially expressed genes as given in \citet{SGT+04}.

Discretization of the data according to \citet{FD81} yielded $K=16$ distinct
gene expression levels. From the $G=102$ genes, we estimated MIs
for  5151 pairs of genes. For each pair, the  mutual information was based on
an estimated $16 \times 16$ contingency table, hence $p=256$.
As  the number of time points is $n=9$, this is a strongly undersampled situation
which requires the use of a regularized estimate of entropy and mutual information.

The distribution of the shrinkage estimates of mutual information 
for all 5151 gene pairs is shown in the left side of \figcite{fig:mihist}.
The right hand side depicts the distribution of mutual information values after
applying the ARACNE procedure, which yields 112 gene pairs with nonzero MIs.

The model selection provided by ARACNE is based on applying the
information processing inequality to all gene triplets. For each triplet, the
gene pair corresponding to the smallest MI is discarded, which has the effect to remove gene-gene links
that correspond to indirect rather than direct interactions. This is similar to a 
procedure used in graphical Gaussian models where correlations are transformed into partial correlations. Thus, both
the ARACNE and the MRNET algorithms can be considered as devices to approximate the conditional
mutual information \citep{MKLB07}. As a result, the 112 nonzero MIs recovered by 
the ARACNE algorithm correspond to statistically detectable direct associations.

The corresponding gene association network is depicted in \figcite{fig:ecolinet}.
The most striking feature of the graph are the ``hubs'' belonging to 
genes hupB, sucA and nuoL. hupB is a well known DNA-binding transcriptional regulator,
whereas both nuoL and sucA are key components of the \emph{E. coli} metabolism.
Note that a Lasso-type procedure (that implicitly limits the number of 
edges that can connect to each node) such as that of \citet{MB06} cannot
recover these hubs.

\section{Discussion}

We proposed a James-Stein-type shrinkage estimator
for inferring entropy and mutual information from small samples.
While this is a challenging problem, we showed that our approach
is highly efficient both statistically and computationally despite its simplicity.

In terms of versatility, our estimator has two distinct advantages over the 
%competing
NSB and Chao-Shen estimators. 
First, in addition to estimating the entropy, it also provides the 
underlying multinomial frequencies for use with the Shannon formula
(\eqcite{eq:shannon}).  This is useful in the context of using mutual information 
to quantify non-linear pairwise dependencies for instance.
Second, unlike NSB, it is a fully analytic estimator.

Hence, our estimator suggests itself for applications in large scale 
estimation problems.  To demonstrate its application in the context of genomics
and systems biology, we have estimated an entropy-based gene dependency network
from expression data in \emph{E. coli}.  This type of approach may prove helpful
to overcome the limitations of linear models currently used in network analysis.

In short, we believe the proposed small-sample entropy estimator will be a 
valuable contribution to the growing toolbox of machine learning and statistics
procedures for high-dimensional data analysis.

%\appendix

%\section{Recipe For Constructing James-Stein-type Shrinkage Estimators}
\section*{Appendix A: Recipe For Constructing James-Stein-type Shrinkage Estimators}

The original James-Stein estimator \citep{JS61} was proposed to estimate
the mean of a multivariate normal distribution from a single ($n=1$!) 
vector observation.   Specifically, if $\bx$  is a sample from $N_p(\bmu, \bI)$
then James-Stein estimator is given by
$$
\hat{\mu}_k^{\text{JS}} = (1- \frac{p-2}{\sum_{k=1}^p x_k^2}) x_k .
$$
Intriguingly, this estimator outperforms the maximum likelihood estimator 
$\hat{\mu}_k^{\text{ML}} = x_k$ in terms of mean squared error 
if the dimension is  $p\geq 3$. Hence, the James-Stein estimator
dominates the maximum likelihood estimator.  

The above estimator can be slightly generalized by shrinking towards the component
average
$\bar x = \sum_{k=1}^p x_k$ rather
than to zero, resulting in
$$
\hat{\mu}_k^{\text{Shrink}} = \hat{\lambda}^\star \bar{x}  + (1- \hat{\lambda}^\star) x_k 
$$
with estimated shrinkage intensity 
$$
\hat{\lambda}^\star=\frac{p-3}{\sum_{k=1}^p (x_k-\bar{x})^2} .
$$

The James-Stein shrinkage principle is very general and can be
put to to use  in many other high-dimensional settings. 
In the following we  summarize a simple 
recipe for constructing James-Stein-type shrinkage estimators along
the lines of \citet{SS05c} and \citet{OS07a}.

In short, there are two key ideas at work in James-Stein shrinkage:~  
\begin{enumerate}
\item[i)] regularization of a high-dimensional estimator $\hbtheta$ by linear
   combination with a lower-dimensional target estimate 
   $\hbtheta^{\text{Target}}$, and
\item[ii)] adaptive estimation of the shrinkage parameter $\lambda$ from 
     the data by quadratic risk minimization.
\end{enumerate}
A general form of a James-Stein-type shrinkage estimator is given by
\begin{equation}
\hbtheta^{\text{Shrink}} = \lambda \hbtheta^{\text{Target}} + (1-\lambda) \hbtheta .
\label{eq:shrink.est}
\end{equation}
Note that $\hbtheta$ and $\hbtheta^{\text{Target}}$ are two very different
estimators (for the same underlying model!). $\hbtheta$ as a high-dimensional estimate with many
independent components has low bias but for small samples a potentially large variance.
In contrast, the target estimate $\hbtheta^{\text{Target}}$ is low-dimensional and therefore
is generally less variable than $\hbtheta$ but at the same time is also more biased. 
The James-Stein estimate is a weighted average of these two estimators,
where the weight is  chosen in a data-driven fashion such that 
$\hbtheta^{\text{Shrink}}$ is improved in terms of mean squared error
relative to \emph{both} $\hbtheta$ and $\hbtheta^{\text{Target}}$.

A key advantage of James-Stein-type shrinkage is that 
the optimal shrinkage intensity $\lambda^\star$ can be calculated
analytically and without knowing the true value $\btheta$, via
\begin{equation}
\lambda^\star = \frac{\sum^p_{k=1} \text{Var}(\hat\theta_k) - 
\text{Cov}(\hat\theta_k, \hat\theta^{\text{Target}}_k) + 
\text{Bias}(\hat\theta_k)\, E(\hat\theta_k-\hat\theta^{\text{Target}}_k) }
                    {\sum^p_{k=1} E[ (\hat\theta_k - \hat\theta^{\text{Target}}_k)^2 ]} . 
\label{eq:opt.lambda}
\end{equation}
A simple estimate of $\lambda^\star$ is obtained by replacing
all variances and covariances in \eqcite{eq:opt.lambda} with their empirical counterparts,
followed by truncation of $\hat\lambda^\star$  at 1 (so that
$\hat\lambda^\star \leq 1$ always holds).

\eqcite{eq:opt.lambda} is discussed in detail in \citet{SS05c} and \citet{OS07a}.
More specialized versions of it are treated, for example, in \citet{LW03} for unbiased $\hbtheta$
and in \citet{Tho68} (unbiased, univariate case with deterministic target).  A very early
version (univariate with zero target) even predates the estimator of James and Stein, see \citet{Goo53}.
For the multinormal setting of \citet{JS61}, \eqcite{eq:shrink.est} and \eqcite{eq:opt.lambda} 
reduce to the shrinkage estimator described in \citet{Sti90}.

James-Stein shrinkage has an empirical Bayes interpretation \citep{EM73}. Note,
however, that only the first two moments of the 
distributions of $\hbtheta^{\text{Target}}$ and $\hbtheta$ need to be specified
in \eqcite{eq:opt.lambda}. Hence,  
James-Stein estimation may be viewed as a \emph{quasi-empirical Bayes} approach 
(in the same sense as in quasi-likelihood, which also requires only the first two moments).

%\section{Computer Implementation}
\section*{Appendix B: Computer Implementation}

The proposed shrinkage estimators of entropy and mutual information, as well as all other investigated
entropy estimators, have been implemented in R \citep{RPROJECT}.
A corresponding R package ``entropy'' was
deposited in the R  archive CRAN and is accessible at the URL 
\url{http://cran.r-project.org/web/packages/entropy/} under the GNU General Public License.

\bibliographystyle{apalike}
\bibliography{preamble,econ,genome,stats,array,sysbio,misc,molevol,med,entropy}

\newcommand{\noopsort}[1]{} \newcommand{\printfirst}[2]{#1}
  \newcommand{\singleletter}[1]{#1} \newcommand{\switchargs}[2]{#2#1}
\begin{thebibliography}{}

\bibitem[Agresti and Hitchcock, 2005]{AH05}
Agresti, A. and Hitchcock, D.~B. (2005).
\newblock {Bayesian} inference for categorical data analysis.
\newblock {\em Statist. Meth. Appl.}, 14:297--330.

\bibitem[Butte et~al., 2000]{BTS+00}
Butte, A.~J., Tamayo, P., Slonim, D., Golub, T.~R., and Kohane, I.~S. (2000).
\newblock Discovering functional relationships between {RNA} expression and
  chemotherapeutic susceptibility using relevance networks.
\newblock {\em Proc. Natl. Acad. Sci. USA}, 97:12182--12186.

\bibitem[Chao and Shen, 2003]{CS03}
Chao, A. and Shen, T.-J. (2003).
\newblock Nonparametric estimation of {Shannon's} index of diversity when there
  are unseen species.
\newblock {\em Environ. Ecol. Stat.}, 10:429--443.

\bibitem[Dobra et~al., 2004]{DHJ+04}
Dobra, A., Hans, C., Jones, B., Nevins, J.~R., Yao, G., and West, M. (2004).
\newblock Sparse graphical models for exploring gene expression data.
\newblock {\em J. Multiv. Anal.}, 90:196--212.

\bibitem[Efron and Morris, 1973]{EM73}
Efron, B. and Morris, C.~N. (1973).
\newblock Stein's estimation rule and its competitors--an empirical {Bayes}
  approach.
\newblock {\em J. Amer. Statist. Assoc.}, 68:117--130.

\bibitem[Fienberg and Holland, 1973]{FH73}
Fienberg, S.~E. and Holland, P.~W. (1973).
\newblock Simultaneous estimation of multinomial cell probabilities.
\newblock {\em J. Amer. Statist. Assoc.}, 68:683--691.

\bibitem[Freedman and Diaconis, 1981]{FD81}
Freedman, D. and Diaconis, P. (1981).
\newblock On the histogram as a density estimator: L2 theory.
\newblock {\em Z. Wahrscheinlichkeitstheorie verw. Gebiete}, 57:453--476.

\bibitem[Friedman, 2004]{Fri04}
Friedman, N. (2004).
\newblock Inferring cellular networks using probabilistic graphical models.
\newblock {\em Science}, 303:799--805.

\bibitem[Geisser, 1984]{Gei84}
Geisser, S. (1984).
\newblock On prior distributions for binary trials.
\newblock {\em The American Statistician}, 38:244--251.

\bibitem[Gelman et~al., 2004]{GCSR04}
Gelman, A., Carlin, J.~B., Stern, H.~S., and Rubin, D.~B. (2004).
\newblock {\em Bayesian Data Analysis}.
\newblock Chapman \& Hall/CRC, Boca Raton, 2nd edition.

\bibitem[Good, 1953]{Good53}
Good, I.~J. (1953).
\newblock The population frequencies of species and the estimation of
  population parameters.
\newblock {\em Biometrika}, 40:237--264.

\bibitem[Goodman, 1953]{Goo53}
Goodman, L.~A. (1953).
\newblock A simple method for improving some estimators.
\newblock {\em Ann. Math. Statist.}, 24:114--117.

\bibitem[Gruber, 1998]{Gru98}
Gruber, M. H.~J. (1998).
\newblock {\em Improving Efficiency By Shrinkage}.
\newblock Marcel Dekker, Inc., New York.

\bibitem[Holste et~al., 1998]{HGH98}
Holste, D., Gro\ss{}e, I., and Herzel, H. (1998).
\newblock {Bayes'} estimators of generalized entropies.
\newblock {\em J. Phys. A: Math. Gen.}, 31:2551--2566.

\bibitem[Horvitz and Thompson, 1952]{HT52}
Horvitz, D.~G. and Thompson, D.~J. (1952).
\newblock A generalization of sampling without replacement from a finite
  universe.
\newblock {\em J. Amer. Statist. Assoc.}, 47:663--685.

\bibitem[James and Stein, 1961]{JS61}
James, W. and Stein, C. (1961).
\newblock Estimation with quadratic loss.
\newblock In {\em Proc. Fourth Berkeley Symp. Math. Statist. Probab.},
  volume~1, pages 361--379, Berkeley. Univ. California Press.

\bibitem[Jeffreys, 1946]{Jef46}
Jeffreys, H. (1946).
\newblock An invariant form for the prior probability in estimation problems.
\newblock {\em Proc. Roc. Soc. (Lond.) A}, 186:453--461.

\bibitem[Kalisch and B\"uhlmann, 2007]{KB07}
Kalisch, M. and B\"uhlmann, P. (2007).
\newblock Estimating high-dimensional directed acyclic graphs with the
  {PC}-algorithm.
\newblock {\em J. Machine Learn. Res.}, 8:613--636.

\bibitem[Krichevsky and Trofimov, 1981]{KT81}
Krichevsky, R.~E. and Trofimov, V.~K. (1981).
\newblock The performance of universal encoding.
\newblock {\em IEEE Trans. Inf. Theory}, 27:199--207.

\bibitem[L\"ahdesm\"aki and Shmulevich, 2008]{LS08}
L\"ahdesm\"aki, H. and Shmulevich, I. (2008).
\newblock Learning the structure of dynamic {Bayesian} networks from time
  series and steady state measurements.
\newblock {\em Mach. Learn.}, 71:185--217.

\bibitem[Ledoit and Wolf, 2003]{LW03}
Ledoit, O. and Wolf, M. (2003).
\newblock Improved estimation of the covariance matrix of stock returns with an
  application to portfolio selection.
\newblock {\em J. Empir. Finance}, 10:603--621.

\bibitem[MacKay, 2003]{Mac03}
MacKay, D. J.~C. (2003).
\newblock {\em Information Theory, Inference, and Learning Algorithms}.
\newblock Cambridge University Press, Cambridge.

\bibitem[Margolin et~al., 2006]{MNB+06}
Margolin, A., Nemenman, I., Basso, K., Wiggins, C., Stolovitzky, G.,
  Dalla~Favera, R., and Califano, A. (2006).
\newblock {ARACNE:} an algorithm for the reconstruction of gene regulatory
  networks in a mammalian cellular context.
\newblock {\em BMC Bioinformatics}, 7 (Suppl. 1):S7.

\bibitem[Meinshausen and B\"uhlmann, 2006]{MB06}
Meinshausen, N. and B\"uhlmann, P. (2006).
\newblock High-dimensional graphs and variable selection with the {Lasso}.
\newblock {\em Ann. Statist.}, 34:1436--1462.

\bibitem[Meyer et~al., 2007]{MKLB07}
Meyer, P.~E., Kontos, K., Lafitte, F., and Bontempi, G. (2007).
\newblock Information-theoretic inference of large transcriptional regulatory
  networks.
\newblock {\em EURASIP J. Bioinf. Sys. Biol.}, page doi:10.1155/2007/79879.

\bibitem[Miller, 1955]{Mil55}
Miller, G.~A. (1955).
\newblock Note on the bias of information estimates.
\newblock In Quastler, H., editor, {\em Information Theory in Psychology II-B},
  pages 95--100. Free Press, Glencoe, IL.

\bibitem[Nemenman et~al., 2002]{NSB02}
Nemenman, I., Shafee, F., and Bialek, W. (2002).
\newblock Entropy and inference, revisited.
\newblock In Dietterich, T.~G., Becker, S., and Ghahramani, Z., editors, {\em
  Advances in Neural Information Processing Systems 14}, pages 471--478,
  Cambridge, MA. MIT Press.

\bibitem[Opgen-Rhein and Strimmer, 2007a]{OS07a}
Opgen-Rhein, R. and Strimmer, K. (2007a).
\newblock Accurate ranking of differentially expressed genes by a
  distribution-free shrinkage approach.
\newblock {\em Statist. Appl. Genet. Mol. Biol.}, 6:9.

\bibitem[Opgen-Rhein and Strimmer, 2007b]{OS07c}
Opgen-Rhein, R. and Strimmer, K. (2007b).
\newblock From correlation to causation networks: a simple approximate learning
  algorithm and its application to high-dimensional plant gene expression data.
\newblock {\em BMC Systems Biology}, 1:37.

\bibitem[Orlitsky et~al., 2003]{OSZ03}
Orlitsky, A., Santhanam, N.~P., and Zhang, J. (2003).
\newblock Always {Good Turing}: asymptotically optimal probability estimation.
\newblock {\em Science}, 302:427--431.

\bibitem[Perks, 1947]{Per47}
Perks, W. (1947).
\newblock Some observations on inverse probability including a new indifference
  rule.
\newblock {\em J. Inst. Actuaries}, 73:285--334.

\bibitem[{R Development Core Team}, 2008]{RPROJECT}
{R Development Core Team} (2008).
\newblock {\em R: A language and environment for statistical computing}.
\newblock R Foundation for Statistical Computing, Vienna, Austria.
\newblock {ISBN} 3-900051-07-0.

\bibitem[Rangel et~al., 2004]{RA+04}
Rangel, C., Angus, J., Ghahramani, Z., Lioumi, M., Sotheran, E., Gaiba, A.,
  Wild, D.~L., and Falciani, F. (2004).
\newblock Modeling {T}-cell activation using gene expression profiling and
  state space modeling.
\newblock {\em Bioinformatics}, 20:1361--1372.

\bibitem[Sch\"afer and Strimmer, 2005a]{SS05a}
Sch\"afer, J. and Strimmer, K. (2005a).
\newblock An empirical {Bayes} approach to inferring large-scale gene
  association networks.
\newblock {\em Bioinformatics}, 21:754--764.

\bibitem[Sch\"afer and Strimmer, 2005b]{SS05c}
Sch\"afer, J. and Strimmer, K. (2005b).
\newblock A shrinkage approach to large-scale covariance matrix estimation and
  implications for functional genomics.
\newblock {\em Statist. Appl. Genet. Mol. Biol.}, 4:32.

\bibitem[Schmidt-Heck et~al., 2004]{SGT+04}
Schmidt-Heck, W., Guthke, R., Toepfer, S., Reischer, H., Duerrschmid, K., and
  Bayer, K. (2004).
\newblock Reverse engineering of the stress response during expression of a
  recombinant protein.
\newblock In ?, editor, {\em Proceedings of the EUNITE symposium, 10-12 June
  2004, Aachen, Germany}, volume~?, pages 407--412, ? Verlag Mainz.

\bibitem[Sch\"urmann and Grassberger, 1996]{SG96}
Sch\"urmann, T. and Grassberger, P. (1996).
\newblock Entropy estimation of symbol sequences.
\newblock {\em Chaos}, 6:414--427.

\bibitem[Stigler, 1990]{Sti90}
Stigler, S.~M. (1990).
\newblock A {Galtonian} perspective on shrinkage estimators.
\newblock {\em Statistical Science}, 5:147--155.

\bibitem[Stinson, 2006]{Stin06}
Stinson, D. (2006).
\newblock {\em {Cryptography: Theory and Practice}}.
\newblock CRC Press.

\bibitem[Strong et~al., 1998]{SK+98}
Strong, S.~P., Koberle, R., {de Ruyter van Steveninck}, R., and Bialek, W.
  (1998).
\newblock Entropy and information in neural spike trains.
\newblock {\em Phys. Rev. Letters}, 80:197--200.

\bibitem[Thompson, 1968]{Tho68}
Thompson, J.~R. (1968).
\newblock Some shrinkage techniques for estimating the mean.
\newblock {\em J. Amer. Statist. Assoc.}, 63:113--122.

\bibitem[Trybula, 1958]{Try58}
Trybula, S. (1958).
\newblock Some problems of simultaneous minimax estimation.
\newblock {\em Ann. Math. Statist.}, 29:245--253.

\bibitem[Tuyl et~al., 2008]{TGM08}
Tuyl, F., Gerlach, R., and Mengersen, K. (2008).
\newblock A comparison of {Bayes}-{Laplace}, {Jeffreys}, and other priors: the
  case of zero events.
\newblock {\em The American Statistician}, 62:40--44.

\bibitem[Vu et~al., 2007]{VYK07}
Vu, V.~Q., Yu, B., and Kass, R.~E. (2007).
\newblock Coverage-adjusted entropy estimation.
\newblock {\em Stat. Med.}, 26:4039--4060.

\bibitem[Yeo and Burge, 2004]{YB04}
Yeo, G. and Burge, C.~B. (2004).
\newblock Maximum entropy modeling of short sequence motifs with applications
  to {RNA} splicing signals.
\newblock {\em J. Comp. Biol.}, 11:377--394.

\end{thebibliography}

\end{document}